\documentclass[11pt,a4paper]{article}
\usepackage[hyperref]{acl2020}
\usepackage{graphicx}
\usepackage{enumitem}
\usepackage{amsfonts}
\usepackage{amsmath}
\usepackage{multirow}
\usepackage{array} 
\usepackage{booktabs}  
\usepackage{amssymb}

\aclfinalcopy

\title{A Survey on Dialog Management: Recent Advances and Challenges}

\author{
  Yinpei Dai$^\dag$, Huihua Yu$^\ddag$, Yixuan Jiang$^\ddag$, Chengguang Tang$^\dag$, Yongbin Li$^\dag$, Jian Sun$^\dag$, \\
    \text{$^\dag$Alibaba Group, Beijing}\\
    \text{$^\ddag$Cornell University}\\
    \texttt{ \{yinpei.dyp,chengguang.tcg,shuide.lyb,jian.sun\}@alibaba-inc.com}
}
\date{}
\begin{document}
\maketitle
\begin{abstract}
    Dialog management (DM) is a crucial component in a
task-oriented dialog system.  Given the dialog history, DM predicts the dialog state and decides the next action that the dialog agent should take. Recently, dialog policy learning has been widely formulated as a Reinforcement Learning (RL) problem, and more works are focus on the applicability of DM. In this paper, we survey recent advances and challenges within three critical topics for DM: (1) improving model scalability to facilitate dialog system modeling in new scenarios, (2) dealing with the data scarcity problem for dialog policy learning, and (3) enhancing the training efficiency  to achieve
better task-completion performance . We believe that this survey\footnote{This survey is translated by Alibaba Cloud International Team based on https://developer.aliyun.com/article/741450.} can shed a light on future research in dialog management.
\end{abstract}

\section{Introduction}
Many efforts have been made to develop highly intelligent human-machine dialog systems since research began on artificial intelligence (AI). Alan Turing proposed the Turing test in 1950\cite{turing2009computing}. He believed that machines could be considered highly intelligent if they passed the Turing test. To pass this test, the machine had to communicate with a real person so that this person believed they were talking to another person. The first-generation dialog systems were mainly rule-based. For example, the ELIZA system\cite{weizenbaum1966eliza} developed by MIT in 1966 was a psychological medical chatbot that matched methods using templates. The flowchart-based dialog system popular in the 1970s simulates state transition in the dialog flow based on the finite state automaton (FSA) model. These machines have transparent internal logic and are easy to analyze and debug. However, they are less flexible and scalable due to their high dependency on expert intervention.

Second-generation dialog systems driven by statistical data (hereinafter referred to as the statistical dialog systems) emerged with the rise of big data technology. At that time, reinforcement learning was widely studied and applied in dialog systems. A representative example is the statistical dialog system based on the Partially Observable Markov Decision Process (POMDP) proposed by Professor Steve Young of Cambridge University in 2005\cite{young2013pomdp}. This system is significantly superior to rule-based dialog systems in terms of robustness. It maintains the state of each round of dialog through Bayesian inference based on speech recognition results and then selects a dialog policy based on the dialog state to generate a natural language response. With a reinforcement learning framework, the POMDP-based dialog system constantly interacts with user simulators or real users to detect errors and optimize the dialog policy accordingly. A statistical dialog system is a modular system not highly dependent on expert intervention. However, it is less scalable, and the model is difficult to maintain.

In recent years, with breakthroughs in deep learning in the image, voice, and text fields, third-generation dialog systems built around deep learning have emerged. These systems still adopt the framework of the statistical dialog systems, but apply a neural network model in each module. Neural network models have powerful representation and language classification and generation capabilities. Therefore, models based on natural language are transformed from generative models, such as Bayesian networks, into deep discriminative models, such as Convolutional Neural Networks (CNNs), Deep Neural Networks (DNNs), and Recurrent Neural Networks (RNNs)\cite{wen2016network}. The dialog state is obtained by directly calculating the maximum conditional probability instead of the Bayesian a posteriori probability. The deep reinforcement learning model is also used to optimize the dialog policy\cite{su2017sample}. In addition, the success of end-to-end sequence-to-sequence technology in machine translation makes end-to-end dialog systems possible. Facebook researchers proposed a task-oriented dialog system based on memory networks\cite{bordes2016learning}, presenting a new way forward in the research of the end-to-end task-oriented dialog systems in third-generation dialog systems. In general, third-generation dialog systems are better than second-generation dialog systems, but a large amount of tagged data is required for effective training. Therefore, improving the cross-domain migration and scalability of the model has become an important area of research.

Common dialog systems are divided into the following three types: chatting systems, task-oriented dialog systems, and QA systems. In a chatting systems, the system generates interesting and informative natural responses to allow human-machine dialog to proceed\cite{serban2017hierarchical}.

In a QA systems, the system analyzes each question and finds a correct answer from its candidate set\cite{berant2013semantic}. A task-oriented dialog (hereinafter referred to as a task dialog) is a task-driven multi-round dialog. The machine determines the user's requirements through understanding, active inquiry, and clarification, makes queries by calling an Application Programming Interface (API), and returns the correct results. Generally, a task dialog is a sequence decision-making process. During the dialog, the machine updates and maintains the internal dialog state by understanding user statements and then selects the optimal action based on the current dialog state, such as determining the requirement, querying restrictions, and providing results.

\begin{figure*}
    \centering
    \includegraphics[width=0.8\textwidth]{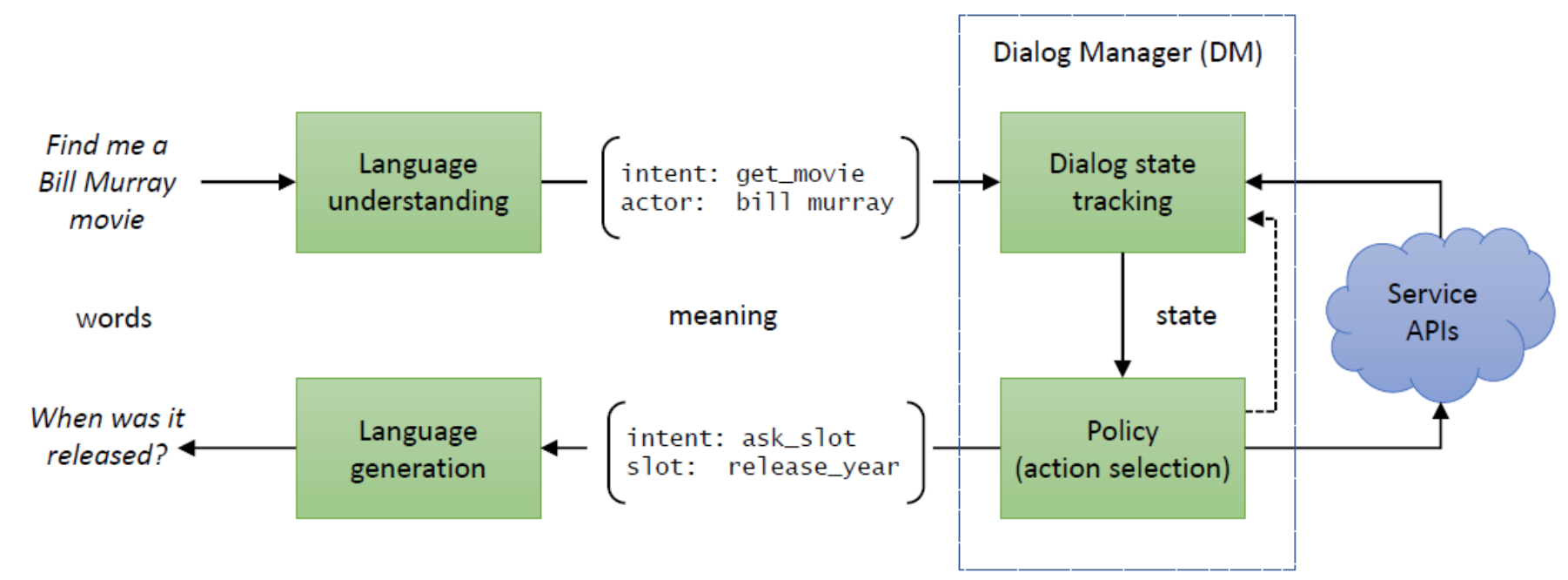}
    \caption{Modular structure of a task-oriented dialog system\cite{gao2018neural}.}
    \label{fig:1}
\end{figure*}

Task-oriented dialog systems are divided by architecture into two categories. One type is a pipeline system that has a modular structure\cite{wen2016network}, as shown in Figure \ref{fig:1}. It consists of four key modules:
\begin{itemize}
    \item Natural Language Understanding (NLU): It identifies and parses a user's text input to obtain semantic tags that can be understood by computers, such as slot-values and intentions.
    \item Dialog State Tracking (DST):  It maintains the current dialog state based on the dialog history. The dialog state is the cumulative meaning of the dialog history, which is generally expressed as slot-value pairs.
    \item Dialog Policy: It outputs the next system action based on the current dialog state. The DST module and the dialog policy module are collectively referred to as the dialog manager (DM).
    \item Natural Language Generation (NLG): It converts system actions to natural language output.
\end{itemize}

This modular system structure is highly interpretable, easy to implement, and applied in most practical task-oriented dialog systems in the industry. However, this structure is not flexible enough. The modules are independent of each other and difficult to optimize together. This makes it difficult to adapt to changing application scenarios. Additionally, due to the accumulation of errors between modules, the upgrade of a single module may require the adjustment of the whole system.

\begin{figure*}
    \centering
    \includegraphics[width=0.8\textwidth]{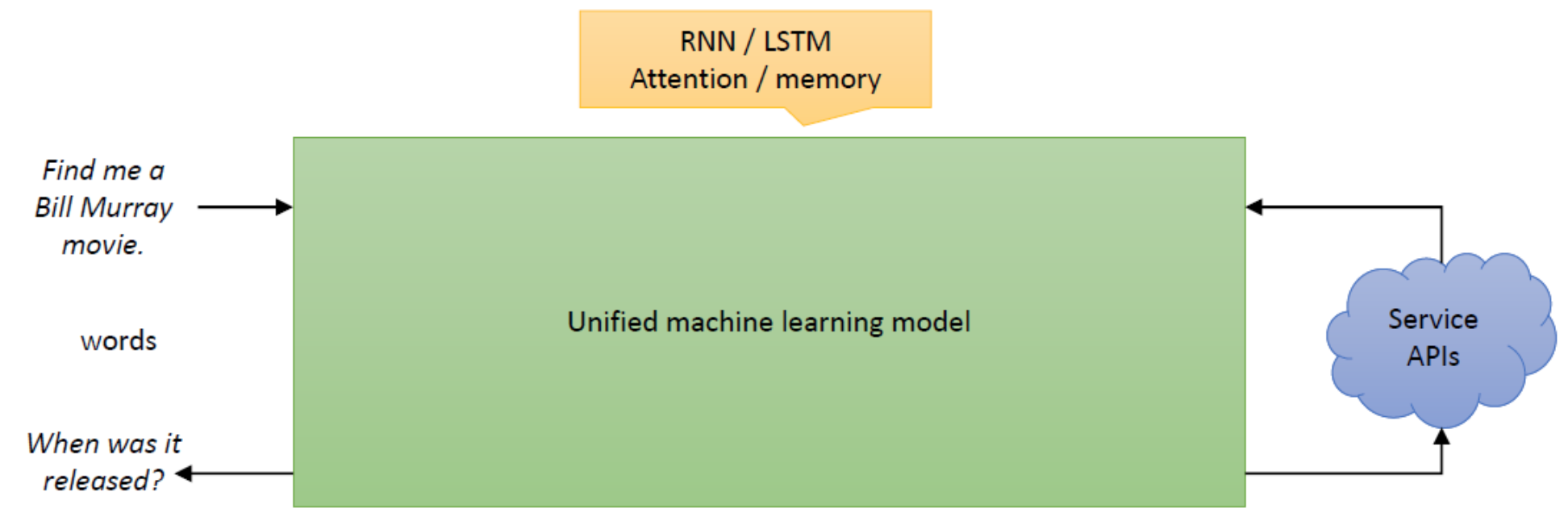}
    \caption{End-to-end structure of a task-oriented dialog system\cite{gao2018neural}}
    \label{fig:2}
\end{figure*}

Another implementation of a task-oriented dialog system is an end-to-end system, which has been a popular field of academic research in recent years\cite{dhingra2016towards, lei2018sequicity, madotto2018mem2seq} (Figure \ref{fig:2}). This type of structure trains an overall mapping relationship from the natural language input on the user side to the natural language output on the machine side. It is highly flexible and scalable, reducing labor costs for design and removing the isolation between modules. However, the end-to-end model places high requirements on the quantity and quality of data and does not provide clear modeling for processes such as slot filling and API calling. This model is still being explored and is as yet rarely applied in the industry.
With higher requirements on product experience, actual dialog scenarios become more complex, and DM needs to be further improved. Traditional DM is usually built in a clear dialog script system (searching for matching answers, querying the user intent, and then ending the dialog) with pre-defined system action space, user intent space, and dialog body. However, due to unpredictable user behaviors, traditional dialog systems are less responsive and have a greater difficulty dealing with undefined situations. In addition, many actual scenarios require cold start without sufficient tagged dialog data, resulting in high data cleansing and tagging costs. DM based on deep reinforcement learning requires a large amount of data for model training. According to the experiments in many academic papers, hundreds of complete sessions are required to train a dialog model, which hinders the rapid development and iteration of dialog systems.

To solve the limitations of traditional DM, researchers in academic and industry circles have begun to focus on how to strengthen the usability of DM. Specifically, they are working to address the following shortcomings in DM:
\begin{itemize}
    \item Poor scalability
    \item Insufficient tagged data
    \item Low training efficiency
\end{itemize}

We will introduce the latest research results in terms of the preceding aspects.

\section{Cutting-Edge Research on Dialog Management}
\subsection{Shortcoming 1: Poor Scalability}

As mentioned above, DM consists of the DST and dialog policy modules. The most representative traditional DST is the neural belief tracker (NBT) proposed by scholars from Cambridge University in 2017\cite{mrkvsic2016neural}. NBT uses neural networks to track the state of complex dialogs in a single domain. By using representation learning, NBT encodes system actions in the previous round, user statements in the current round, and candidate slot-value pairs to calculate semantic similarity in a high dimensional space and detect the slot value output by the user in the current round. Therefore, NBT can identify slot values that are not in the training set but semantically similar to those in the set by using the word vector expression of the slot-value pair. This avoids the need to create a semantic dictionary. As such, the slot values can be extended. Later, Cambridge scholars further improved NBT\cite{ramadan2018large, weisz2018sample} by changing the input slot-value pair to the domain-slot-value triple. The recognition results of each round are accumulated using model learning instead of manual rules. All data is trained by the same model. Knowledge is shared among different domains, leaving the total number of parameters unchanged as the number of domains increases. Among traditional dialog policy research, the most representative is the ACER-based policy optimization proposed by Cambridge scholars\cite{su2017sample, weisz2018sample}.

By applying the experience replay technique, the authors tried both the trust region actor-critic model and the episodic natural actor-critic model. The results proved that the deep AC-based reinforcement learning algorithms were the best in sample utilization efficiency, algorithm convergence, and dialog success rate.

However, traditional DM still needs to be improved in terms of scalability, specifically in the following three respects:
\begin{enumerate}
    \item How to deal with changing user intents.
    \item How to deal with changing slots and values.
    \item How to deal with changing system actions.
\end{enumerate}

\subsubsection{Changing User Intents}
\begin{figure}
    \centering
    \includegraphics[width=0.5\textwidth]{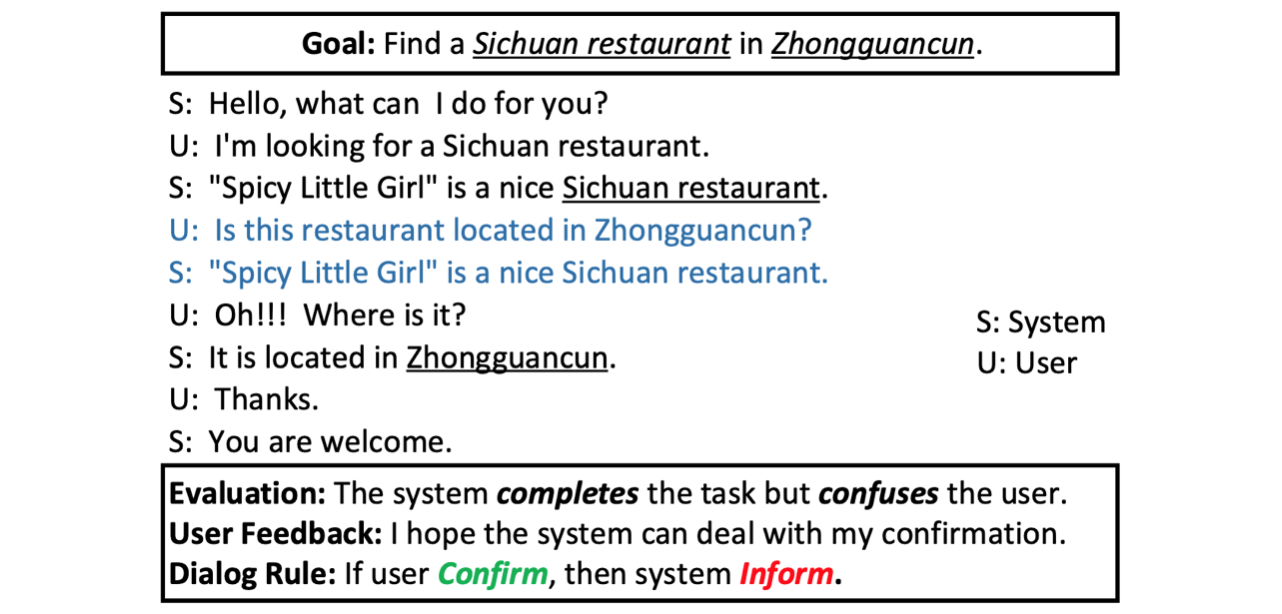}
    \caption{Example of a dialog with new intent\cite{wang2018teacher}}
    \label{fig:3}
\end{figure}

If a system does not take the user intent into account, it will often provide nonsensical answers. As shown in Figure \ref{fig:3}, the user's ``confirm" intent is not considered. A new dialog script must be added to help the system deal with this problem.

\begin{figure}
    \centering
    \includegraphics[width=0.5\textwidth]{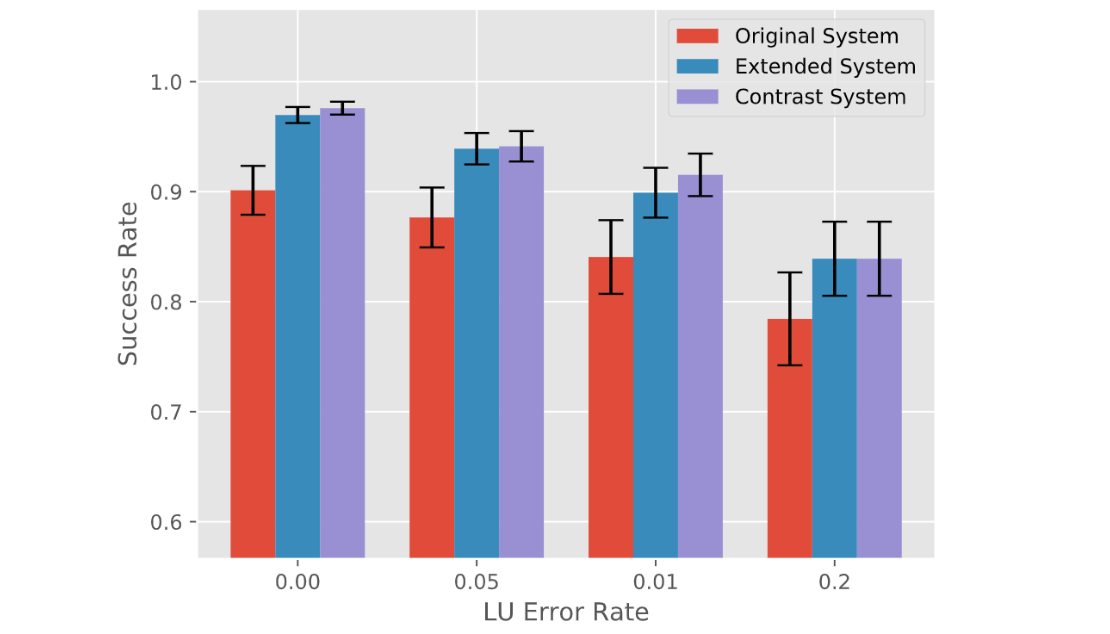}
    \caption{Comparison of various models at different noise levels}
    \label{fig:4}
\end{figure}

The traditional model outputs a fixed one-hot vector of the old intent category. Once a new user intent not in the training set appears, vectors need to be changed to include the new intent category, and the new model needs to be retrained. This makes the model less maintainable and scalable. \citet{wang2018teacher} proposes a teacher-student learning framework to solve this problem. In the teacher-student training architecture, the old model and logical rules for new user intents are used as the teacher, and the new model as a student. This architecture uses knowledge distillation technology. Specifically, for the old intent set, the probability output of the old model directly guides the training of the new model. For the new intent, the logical rules are used as new tagged data to train the new model. In this way, the new model no longer needs to interact with the environment for re-training. The paper presented the results of an experiment performed on the DSTC2 dataset. The confirm intent is deliberately removed and then added as a new intent to the dialog body to verify whether the new model is adaptable. Figure \ref{fig:4} shows the experiment result. The new model (Extended System), the model containing all intents (Contrast System), and the old model are compared. The result shows that the new model achieves satisfactory success rates in extended new intent identification at different noise levels.

Of course, systems with this architecture need to be further trained. CDSSM, a semantic similarity matching model proposed in \cite{chen2016zero}, can identify extended user intents without tagged data and model re-training. Based on the natural description of user intents in the training set, CDSSM directly learns an intent embedding encoder and embeds the description of any intent into a high dimensional semantic space. In this way, the model directly generates corresponding intent embedding based on the natural description of the new intent and then identifies the intent. Many models that improve scalability mentioned below are designed with similar ideas. Tags are moved from the output end of the model to the input end, and neural networks are used to perform semantic encoding on tags (tag names or natural descriptions of the tags) to obtain certain semantic vectors and then match their semantic similarity.

The work of \cite{rajendran2019learning} provides another idea. Through man-machine collaboration, manual customer services are used to deal with user intents not in the training set after the system is launched. This model uses an additional neural parser to determine whether manual customer service is required based on the dialog state vector extracted from the current model. If it is, the model distributes the current dialog to online customer service. If not, the model makes a prediction. The parser obtained through data learning can determine whether the current dialog contains a new intent, and responses from customer service are regarded as correct by default. This man-machine collaboration mechanism effectively deals with user intents not found in the training set during online testing and significantly improves the accuracy of the dialog.

\subsubsection{Changing Slots and Slot Values}
\begin{figure}
    \centering
    \includegraphics[width=0.5\textwidth]{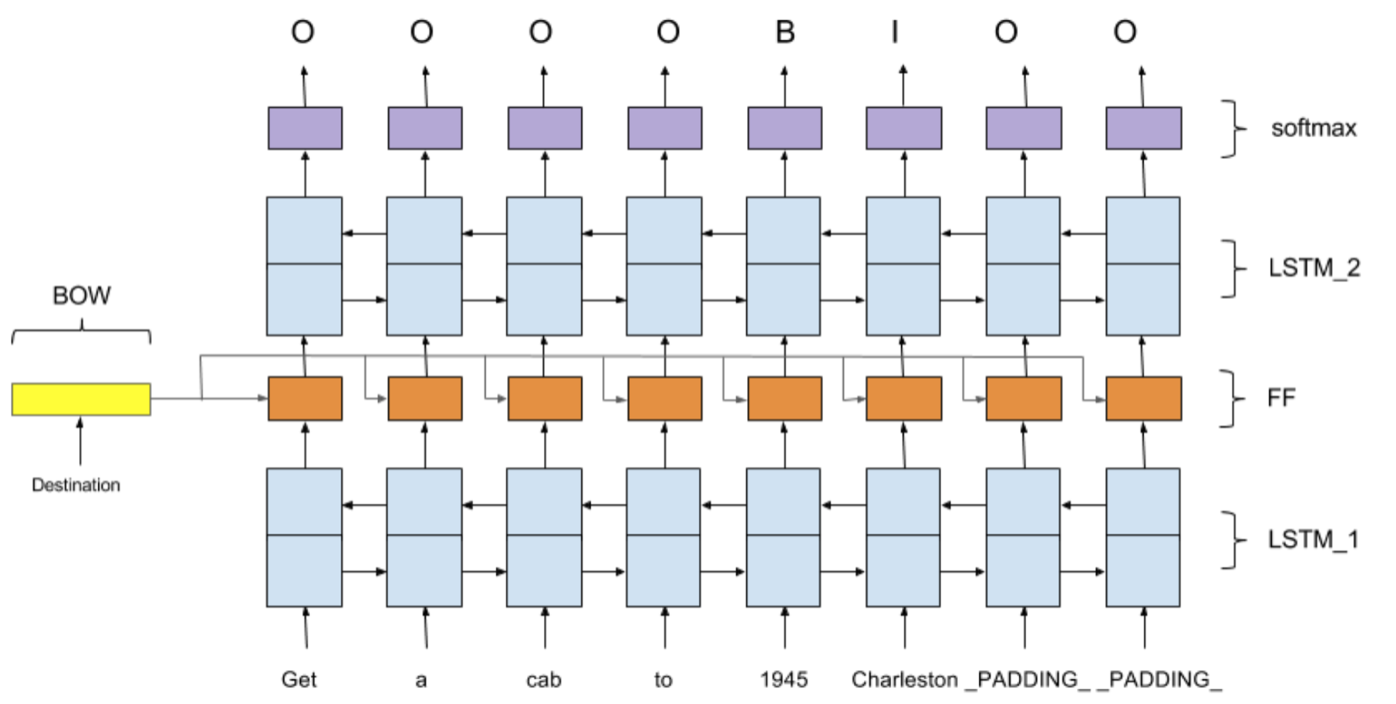}
    \caption{Concept tagger structure}
    \label{fig:5}
\end{figure}
In dialog state tracking involving multiple or complex domains, dealing with changing slots and slot values has always been a challenge. Some slots have non-enumerative slot values, for example, the time, location, and user name. Their slot value sets, such as flights or movie theater schedules, change dynamically. In traditional DST, the slot and slot value set remain unchanged by default, which greatly reduces the system scalability.

Google researchers\cite{rastogi2017scalable} proposed a candidate set for slots with non-enumerative slot values. A candidate set is maintained for each slot. The candidate set contains a maximum of k possible slot values in the dialog and assigns a score to each slot value to indicate the user's preference for the slot value in the current dialog. The system uses a two-way RNN model to find the value of a slot in the current user statement and then score and re-rank it with existing slot values in the candidate set. In this way, the DST of each round only needs to make a judgment on a limited slot value set, allowing us to track non-enumerative slot values. To track slot values not in the set, we can use a sequence tagging model\cite{mesnil2013investigation} or a semantic similarity matching model such as the neural belief tracker\cite{mrkvsic2016neural}. Some work \cite{dai2018elastic, weld2021survey} also uses CRFs for open-ontology slot-filling.

The preceding are solutions for non-fixed slot values, but what about changing slots in the dialog body? In the work of \cite{bapna2017towards}, a slot description encoder is used to encode the natural language description of existing and new slots. The obtained semantic vectors representing the slot are sent with user statements as inputs to the Bi-LSTM model, and the identified slot values are output as sequence tags, as shown in Figure \ref{fig:5}. The paper makes an acceptable assumption that the natural language description of any slot is easy to obtain. Therefore, a concept tagger applicable to multiple domains is designed, and the slot description encoder is simply implemented by the sum of simple word vectors. Experiments show that this model can quickly adapt to new slots. Compared with the traditional method, this method greatly improves scalability. Futher work \cite{liu2020coach, shah2019robust} have been proposed based on the concept tagger for zero-shot cross-domain slot-filling.

\begin{figure*}
    \centering
    \includegraphics[width=0.8\textwidth]{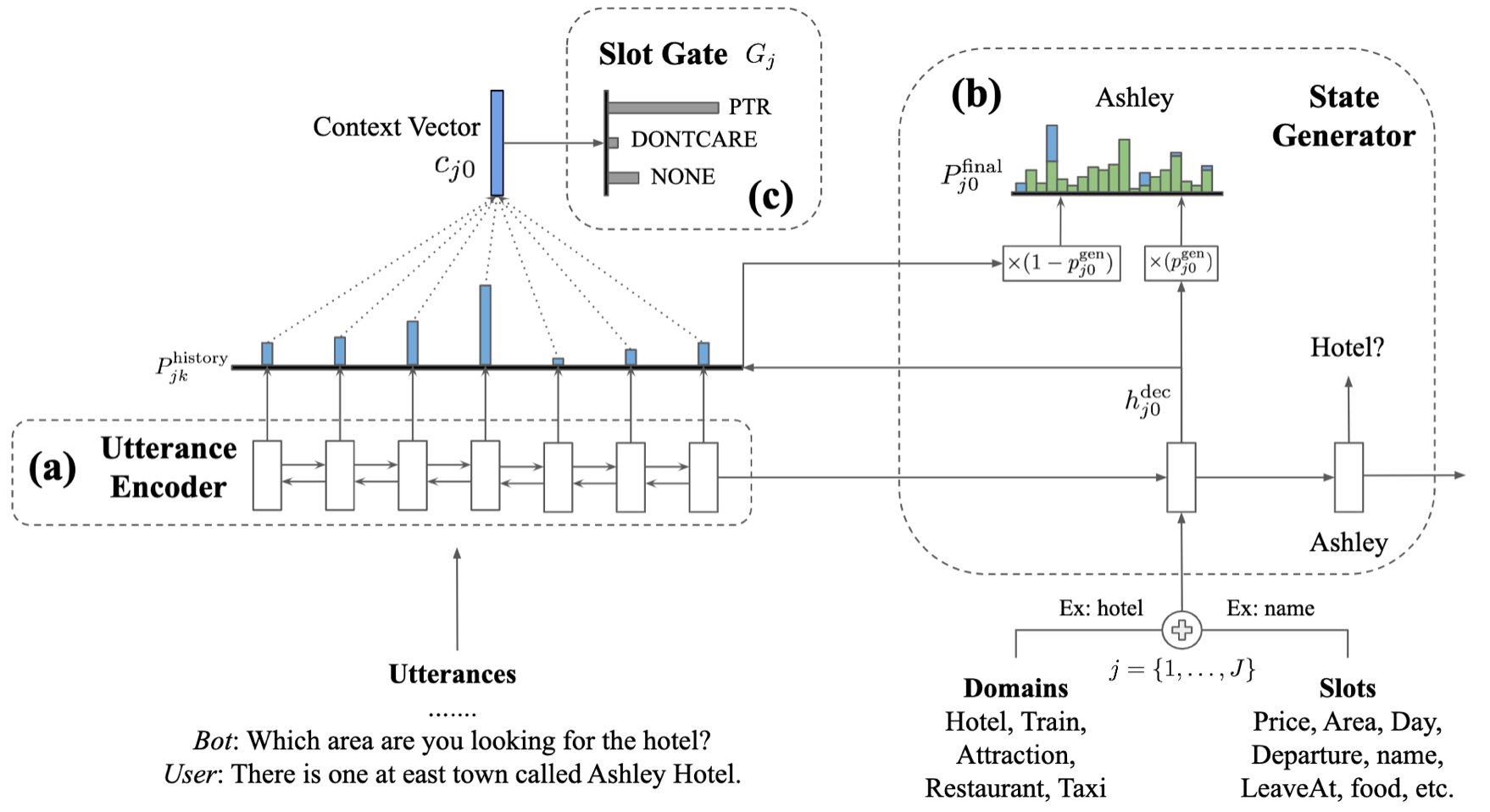}
    \caption{TRADE model framework}
    \label{fig:6}
\end{figure*}

With the development of sequence-to-sequence technology in recent years, many researchers are looking at ways to use the end-to-end neural network model to generate the DST results as a sequence. Common techniques such as attention mechanisms and copy mechanisms are used to improve the generation effect. In the famous MultiWOZ dataset for multi-domain dialogs, the team led by Professor Pascale Fung from Hong Kong University of Science and Technology used the copy network to significantly improve the recognition accuracy of non-enumerative slot values\cite{wu2019transferable}. Figure \ref{fig:6} shows the TRADE model proposed by the team. Each time the slot value is detected, the model performs semantic encoding for different combinations of domains and slots and uses the result as the initial position input of the RNN decoder. The decoder directly generates the slot value through the copy network. In this way, both non-enumerative slot values and changing slot values can be generated by the same model. Therefore, slot values can be shared between domains, allowing the model to be widely used. Built upon the recent powerful transformers, many work \cite{mehri2021gensf, lee2021dialogue} propose to use GPT for slot-filling generation.

Recent research tends to view multi-domain DST as a machine reading and understanding task and transform generative models such as TRADE into discriminative models\cite{zhou2019multi, zhang-etal-2020-find, yucross}. Non-enumerative slot values are tracked by a machine reading and understanding task like SQuAD\cite{rajpurkar-etal-2018-know}, in which the text span in the dialog history and questions is used as the slot value. Enumerative slot values are tracked by a multi-choice machine reading and understanding task, in which the correct value is selected from the candidate values as the predicted slot value. By combining deep context words such as ELMO and BERT, these new models obtain the optimal results from the MultiWOZ dataset. The work \cite{henderson2020convex} from PolyAI also leverage pre-trained slot-filling for few-shot adaptation.

\subsubsection{Changing System Actions}

\begin{figure}
    \centering
    \includegraphics[width=0.5\textwidth]{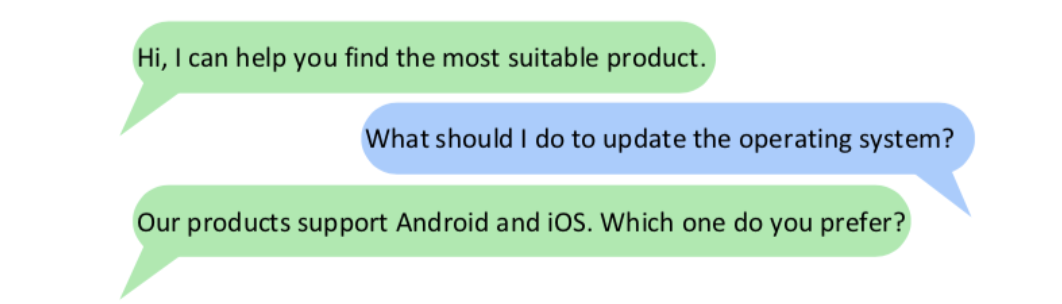}
    \caption{Example of a dialog where the dialog system encounters an undefined system action\cite{wang2019incremental}}
    \label{fig:7}
\end{figure}

The last factor affecting scalability is the difficulty of pre-defining the system action space. As shown in Figure \ref{fig:7}, when designing an electronic product recommendation system, you may ignore questions like how to upgrade the product operating system, but you cannot stop users from asking questions the system cannot answer. If the system action space is pre-defined, irrelevant answers may be provided to questions that have not been defined, greatly compromising the user experience.

\begin{figure*}
    \centering
    \includegraphics[width=1.0\textwidth]{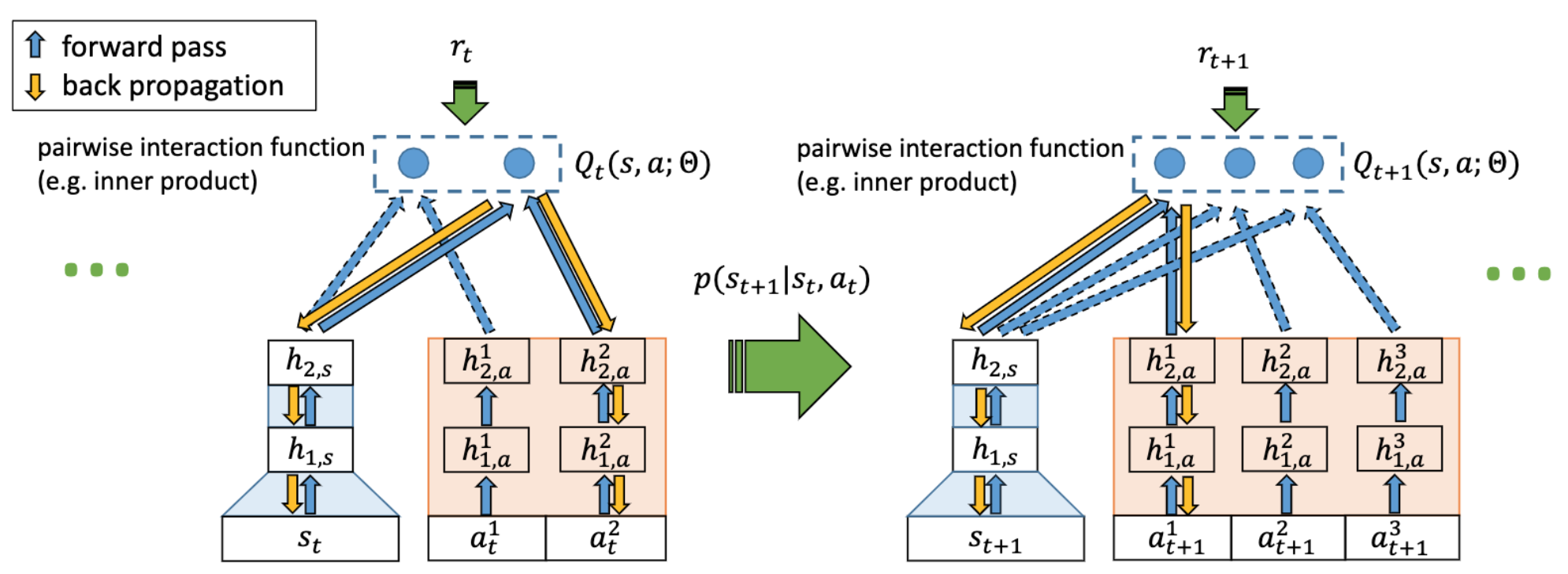}
    \caption{DRRN model, in which round t has two candidate actions, and round t+1 has three candidate actions}
    \label{fig:8}
\end{figure*}

In this case, we need to design a dialog policy network that helps the system quickly expand its actions. The first attempt to do this was made by Microsoft\cite{he2015deep}, who modifies the classic DQN structure to enable reinforcement learning in an unrestricted action space. The dialog task in this paper is a text game mission task. Each round of action is a single sentence, with an uncertain number of actions. The story varies with the action. The author proposed a new model, Deep Reinforcement Relevance Network (DRRN), which matches the current dialog state with optional system actions by semantic similarity matching to obtain the Q function. Specifically, in a round of dialog, each action text of an uncertain length is encoded by a neural network to obtain a system action vector with a fixed length. The story background text is encoded by another neural network to obtain a dialog state vector with a fixed length. The two vectors are used to generate the final Q value through an interactive function, such as dot product. Figure \ref{fig:8} shows the structure of the model designed in the paper. Experiments show that DRRN outperforms traditional DQN (using the padding technique) in the text games ``Saving John" and ``Machine of Death".

\begin{figure}
    \centering
    \includegraphics[width=0.5\textwidth]{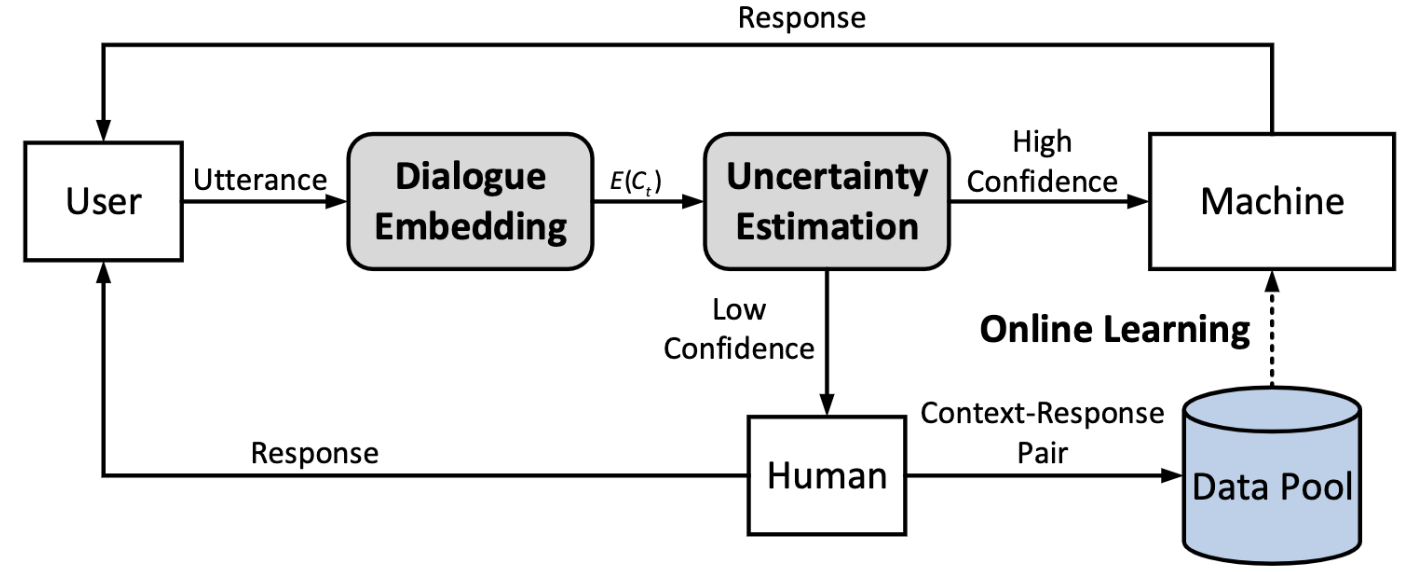}
    \caption{The Overall framework of IDS}
    \label{fig:9}
\end{figure}

In another paper\cite{wang2019incremental}, the author wanted to solve this problem from the perspective of the entire dialogue system and proposed the Incremental Dialogue System (IDS), as shown in Figure \ref{fig:9}. IDS first encodes the dialog history to obtain the context vector through the Dialog Embedding module and then uses a VAE-based Uncertainty Estimation module to evaluate, based on the context vector, the confidence level used to indicate whether the current system can give correct answers. Similar to active learning, if the confidence level is higher than the threshold, DM scores all available actions and then predicts the probability distribution based on the softmax function. If the confidence level is lower than the threshold, the tagger is requested to tag the response of the current round (select the correct response or create a new response). The new data obtained in this way is added to the data pool to update the model online. With this human-teaching method, IDS not only supports learning in an unrestricted action space, but also quickly collects high-quality data, which is quite suitable for actual production.

\subsection{Shortcoming 2: Insufficient Tagged Data}

The extensive application of dialog systems results in diversified data requirements. To train a task-oriented dialog system, as much domain-specific data as possible is needed, but quality tagged data is costly. Scholars have tried to solve this problem in several ways: (1) using machines to tag data to reduce the tagging costs; (2) mining the dialog structure to use non-tagged data efficiently; (3) optimizing the data collection policy to efficiently obtain high-quality data; (4) using meta-learning approaches \cite{qian2019domain, dai2020learning,xu2020meta, dingliwal2021few}; and (5) using pre-training dialog models \cite{hosseini2020simple,peng2020soloist,wu2020tod, dai2021preview}. In this survey we mainly discuss the first three kinds, and leave the last two topics in the future. 

\subsubsection{Automatic Tagging}

\begin{figure*}
    \centering
    \includegraphics[width=1.0\textwidth]{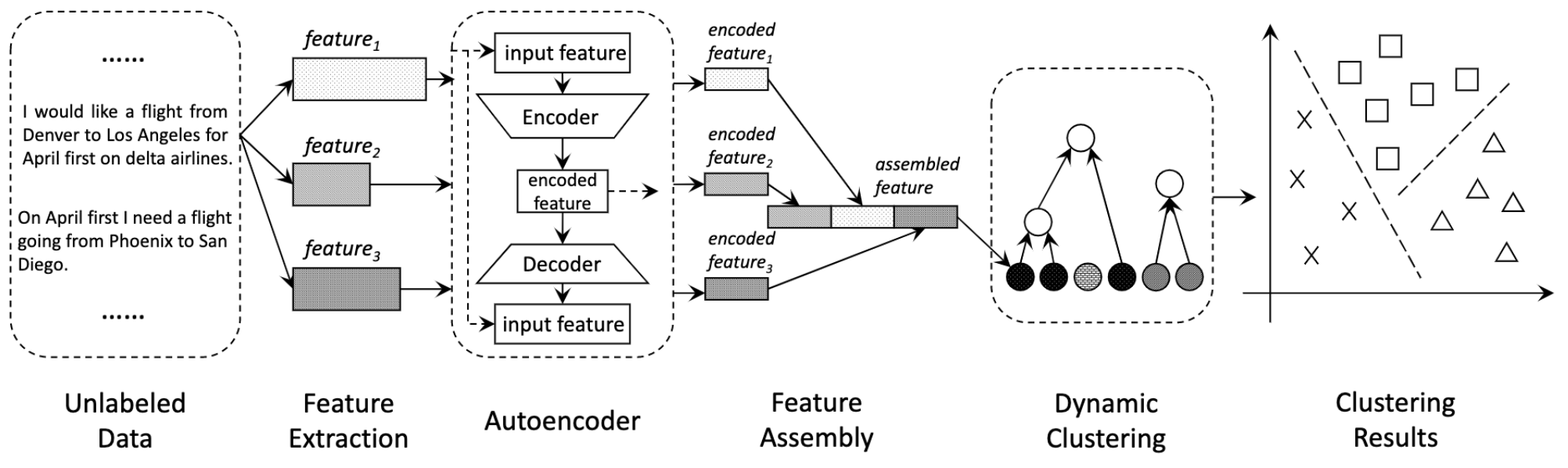}
    \caption{Auto-dialabel model}
    \label{fig:10}
\end{figure*}

To address the cost and inefficiency of manual tagging, scholars hope to use supervised learning and unsupervised learning to allow machines to assist in manual tagging. \citet{shi2018auto} proposed the auto-dielabel architecture, which automatically groups intents and slots in the dialog data by using the unsupervised learning method of hierarchical clustering to automatically tag the dialog data (the specific tag of the category needs to be manually determined). This method is based on the assumption that expressions of the same intent may share similar background features. Initial features extracted by the model include word vectors, part-of-speech (POS) tags, noun word clusters, and Latent Dirichlet allocation (LDA). All features are encoded by the auto-encoder into vectors of the same dimension and spliced. Then, the inter-class distance calculated by the radial bias function (RBF) is used for dynamic hierarchical clustering. Classes that are closest to each other are merged automatically until the inter-class distance between the classes is greater than the threshold. Figure \ref{fig:10} shows the model framework.

In another paper\cite{haponchyk2018supervised}, supervised clustering is used to implement machine tagging. The author views each dialog data record as a graph node and sees the clustering process as the process of identifying the minimum spanning forest. The model uses a support vector machine (SVM) to train the distance scoring model between nodes in the QA dataset through supervised learning. It then uses the structured model and the minimum subtree spanning algorithm to derive the class information corresponding to the dialog data as the hidden variable. It generates the best cluster structure to represent the user intent type.

\subsubsection{Dialog Structure Mining}

Due to the lack of high-quality tagged data for training dialog systems, finding ways to fully mine implicit dialog structures or information in the untagged dialog data has become a popular area of research. Implicit dialog structures or information contribute to the design of dialog policies and the training of dialog models to some extent.

\begin{figure*}
    \centering
    \includegraphics[width=0.7\textwidth]{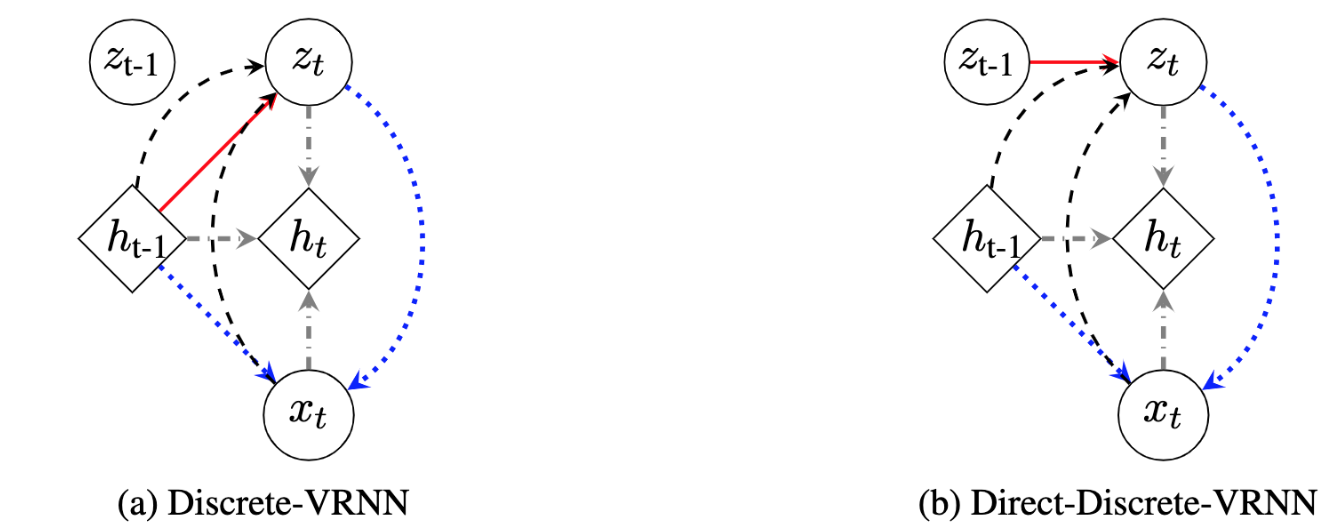}
    \caption{D-VRNN and DD-VRNN}
    \label{fig:11}
\end{figure*}

One paper\cite{shi2019unsupervised} proposed to use unsupervised learning in a variational RNN (VRNN) to automatically learn hidden structures in dialog data. The author provides two models that can obtain the dynamic information in a dialog: Discrete-VRNN (D-VRNN) and Direct-Discrete-VRNN (DD-VRNN). As shown in Figure \ref{fig:11}, $x_t$ indicates the $t-$th round of dialog, $h_t$ indicates the hidden variable of the dialog history, and $z_t$ indicates the hidden variable (one-dimensional one-hot discrete variable) of the dialog structure. The difference between the two models is that for D-VRNN, the hidden variable $z_t$ depends on $h_{t-1}$, while for DD-VRNN, the hidden variable $z_t$ depends on $z_{t-1}$. Based on the maximum likelihood of the entire dialog, VRNN uses some common methods of VAE to estimate the distribution of a posteriori probabilities of the hidden variable $z_t$.

\begin{figure}
    \centering
    \includegraphics[width=0.5\textwidth]{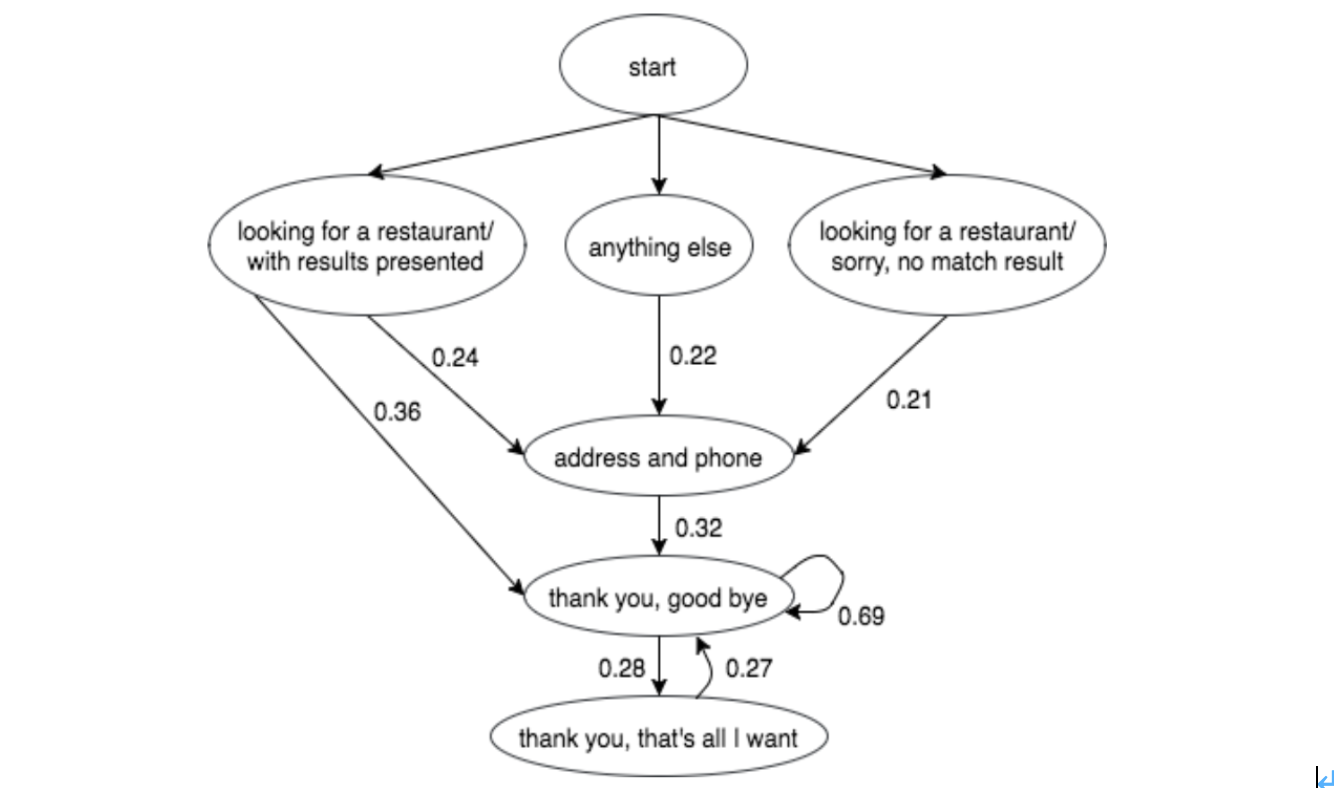}
    \caption{Dialog stream structure mined by D-VRNN from the dialog data related to restaurants}
    \label{fig:12}
\end{figure}

The experiments in the paper show that VRNN is superior to the traditional HMM method. VRNN also adds the dialog structure information to the reward function, supporting faster convergence of the reinforcement learning model. Figure \ref{fig:12} shows the transition probability of the hidden variable $z_t$ in restaurants mined by D-VRNN. There are also some future work to use structured attention network\cite{qiu2020structured} and graph neural network \cite{sun2021unsupervised} to extract hidden dialog structures. 

\begin{figure*}
    \centering
    \includegraphics[width=1.0\textwidth]{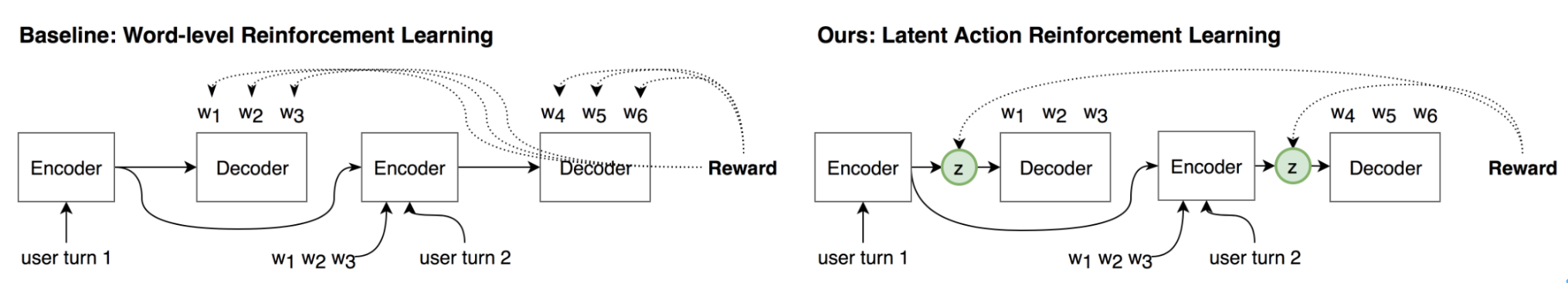}
    \caption{Baseline model and latent action model}
    \label{fig:13}
\end{figure*}

CMU scholars\cite{zhao2019rethinking} also tried to use the VAE method to deduce system actions as hidden variables and directly use them for dialog policy selection. This can alleviate the problems caused by insufficient predefined system actions. As shown in Figure \ref{fig:13}, for simplicity, an end-to-end dialog system framework is used in the paper. The baseline model is an RL model at the word level (that is, a dialog action is a word in the vocabulary). The model uses an encoder to encode the dialog history and then uses a decoder to decode it and generate a response. The reward function directly compares the generated response statement with the real response statement. Compared with the baseline model, the latent action model adds a posterior probability inference between the encoder and the decoder and uses discrete hidden variables to represent the dialog actions without any manual intervention. The experiment shows that the end-to-end RL model based on latent actions is superior to the baseline model in terms of statement generation diversity and task completion rate.

\subsubsection{Data Collection}

\begin{figure*}
    \centering
    \includegraphics[width=0.8\textwidth]{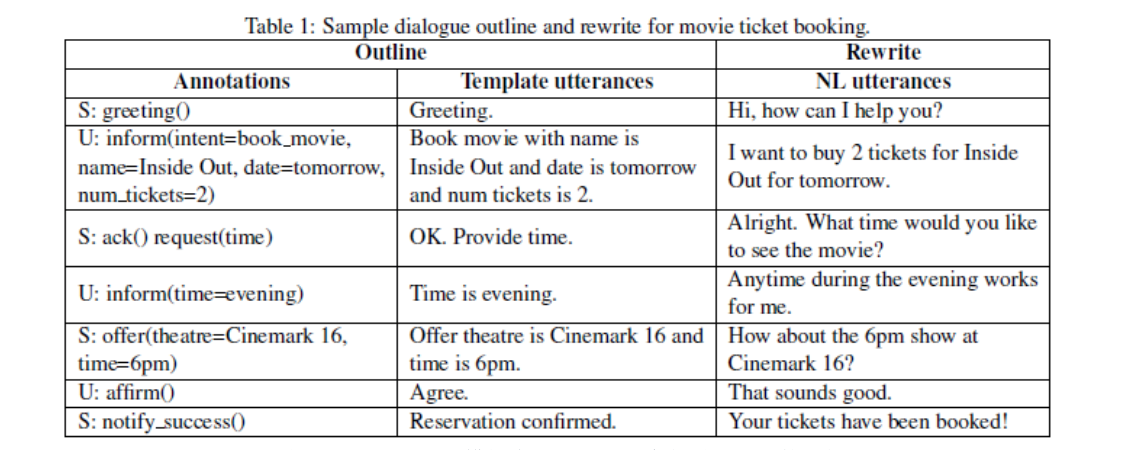}
    \caption{Examples of dialog outline, template-based dialog generation, and crowdsourcing-based dialog rewrite}
    \label{fig:14}
\end{figure*}

Recently, Google researchers proposed a method to quickly collect dialog data\cite{shah2018bootstrapping} (see Figure \ref{fig:14}): First, use two rule-based simulators to interact to generate a dialog outline, which is a dialog flow framework represented by semantic tags. Then, convert the semantic tags into natural language dialogs based on templates. Finally, rewrite the natural statements by crowdsourcing to enrich the language expressions of dialog data. This reverse data collection method features high collection efficiency and complete and highly available data tags, reducing the cost and workload of data collection and processing.

This method is a machine-to-machine (M2M) data collection policy, in which a wide range of semantic tags for dialog data are generated, and then crowd-source to generate a large number of dialog utterances. However, the generated dialogs cannot cover all the possibilities in real scenarios. In addition, the effect depends on the simulator \cite{li2016user,liu2017iterative, chen2020schema,tseng2021transferable}.

In relevant academic circles, two other methods are commonly used to collect data from dialog systems:  human-to-machine (H2M)  and  human-to-human (H2H) . The H2H method requires a multi-round dialog between the user, played by a crowdsourced staff member, and the customer service personnel, played by another crowdsourced staff member. The user proposes requirements based on specified dialog targets such as buying an airplane ticket, and the customer service staff annotates the dialog tags and makes responses. This mode is called the Wizard-of-Oz framework. Many dialog datasets, such as WOZ\cite{wen2016network} and MultiWOZ \cite{budzianowski2018multiwoz}, are collected in this mode. The H2H method helps us get dialog data that is the most similar to that of actual service scenarios. However, it is costly to design different interactive interfaces for different tasks and to clean up incorrect annotations. The H2M data collection policy allows users and trained machines to interact with each other. This way, we can directly collect data online and continuously improve the DM model through RL. The famous DSTC2\&3 dataset was collected in this way. The performance of the H2M method depends largely on the initial performance of the DM model. In addition, the data collected online has a great deal of noise, which results in high clean-up costs and affects the model optimization efficiency.

\subsection{Shortcoming 3: Low Training Efficiency}

With the successful application of deep RL in the Go game, this method is also widely used in the task dialog systems. For example, the ACER dialog management method in one paper\cite{su2017sample} combines model-free deep RL with other techniques such as Experience Replay, belief domain constraints, and pre-training. This greatly improves the training efficiency and stability of RL algorithms in task dialog systems.

However, simply applying the RL algorithm cannot meet the actual requirements of dialog systems. One reason is that dialogs lack clear rules, reward functions, simple and clear action spaces, and perfect environment simulators that can generate hundreds of millions of quality interactive data records. Dialog tasks include changing slot values, actions, and intents, which significantly increases the action space of the dialog system and makes it difficult to define. When traditional flat RL methods are used, the curse of dimensionality may occur due to one-hot encoding of all system actions. Therefore, these methods are no longer suitable for handling complex dialogs with large action spaces. For this reason, scholars have tried many other methods, including model-free RL, model-based RL, and human-in-the-loop.

\subsubsection{Model-Free RL - HRL}

Hierarchical Reinforcement Learning (HRL) divides a complex task into multiple sub-tasks to avoid the curse of dimensionality in traditional flat RL methods. In one paper\cite{peng2017composite}, HRL was applied to task dialog systems for the first time. The authors divided a complex dialog task into multiple sub-tasks by time. For example, a complex travel task can be divided into sub-tasks, such as booking tickets, booking hotels, and renting cars. Accordingly, they designed a dialog policy network of two layers. One layer selects and arranges all sub-tasks, and the other layer executes specific sub-tasks.

\begin{figure}
    \centering
    \includegraphics[width=0.5\textwidth]{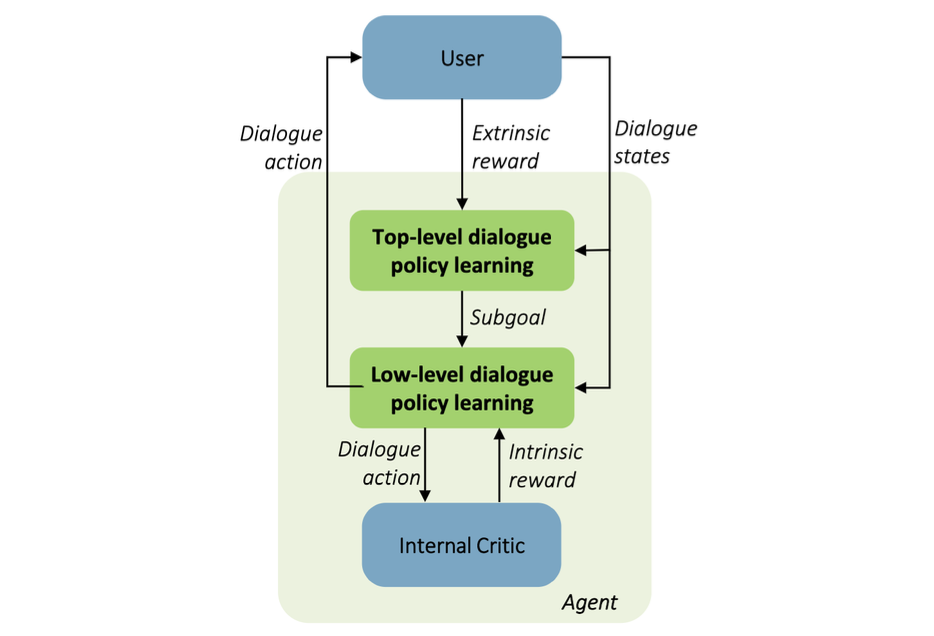}
    \caption{The HRL framework of a task-oriented dialog system}
    \label{fig:15}
\end{figure}

The DM model they proposed consists of two parts, as shown in Figure \ref{fig:15}:
\begin{itemize}
    \item Top-level policy:  Selects a sub-task based on the dialog state.
    \item Low-level policy:  Completes a specific dialog action in a sub-task.
\end{itemize}
The global dialog state tracker records the overall dialog state. After the entire dialog task is completed, the top-level policy receives an external reward.

The model also has an internal critic module to estimate the possibility of completing the sub-tasks (the degree of slot filling for sub-tasks) based on the dialog state. The low-level policy receives an intrinsic reward from the internal critic module based on the degree of completion of the sub-task.

\begin{figure}
    \centering
    \includegraphics[width=0.5\textwidth]{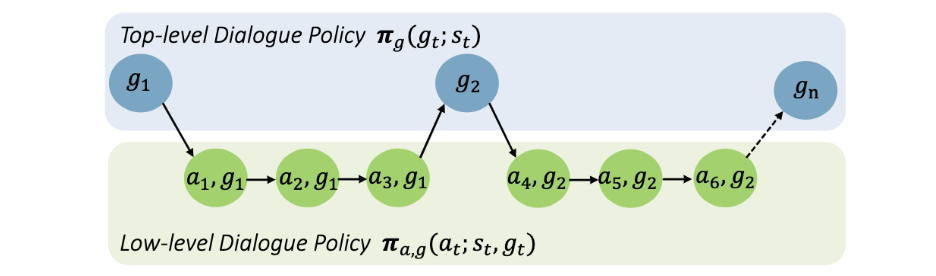}
    \caption{Policy selection process of HRL}
    \label{fig:16}
\end{figure}

For complex dialogs, a basic system action is selected at each step of traditional RL methods, such as querying the slot value or confirming constraints. In the HRL mode, a set of basic actions is selected based on the top-level policy, and then a basic action is selected from the current set based on the low-level policy, as shown in Figure \ref{fig:16}. This hierarchical division of action spaces covers the time sequence constraints between different sub-tasks, which facilitates the completion of composite tasks. In addition, the intrinsic reward effectively relieves the problem of sparse rewards, accelerating RL training, preventing frequent switching of the dialog between different sub-tasks, and improving the accuracy of action prediction. Of course, the hierarchical design of actions requires expert knowledge, and the types of sub-tasks need to be determined by experts. Recently, tools that can automatically discover dialog sub-tasks have appeared\cite{kristianto2018autonomous, tang2018subgoal}. By using unsupervised learning methods, these tools automatically split the dialog state sequence of the whole dialog history, without the need to manually build a dialog sub-task structure.

\subsubsection{Model-free RL - FRL}

Feudal Reinforcement Learning (FRL) is a suitable solution to large dimension issues. HRL divides a dialog policy into sub-policies based on different task stages in the time dimension, which reduces the complexity of policy learning. FRL divides a policy in the space dimension to restrict the action range of each sub-policy, which reduces the complexity of sub-policies. FRL does not divide a task into sub-tasks. Instead, it uses the abstract functions of the state space to extract useful features from dialog states. Such abstraction allows FRL to be applied and migrated between different domains, achieving high scalability.

\begin{figure*}
    \centering
    \includegraphics[width=0.7\textwidth]{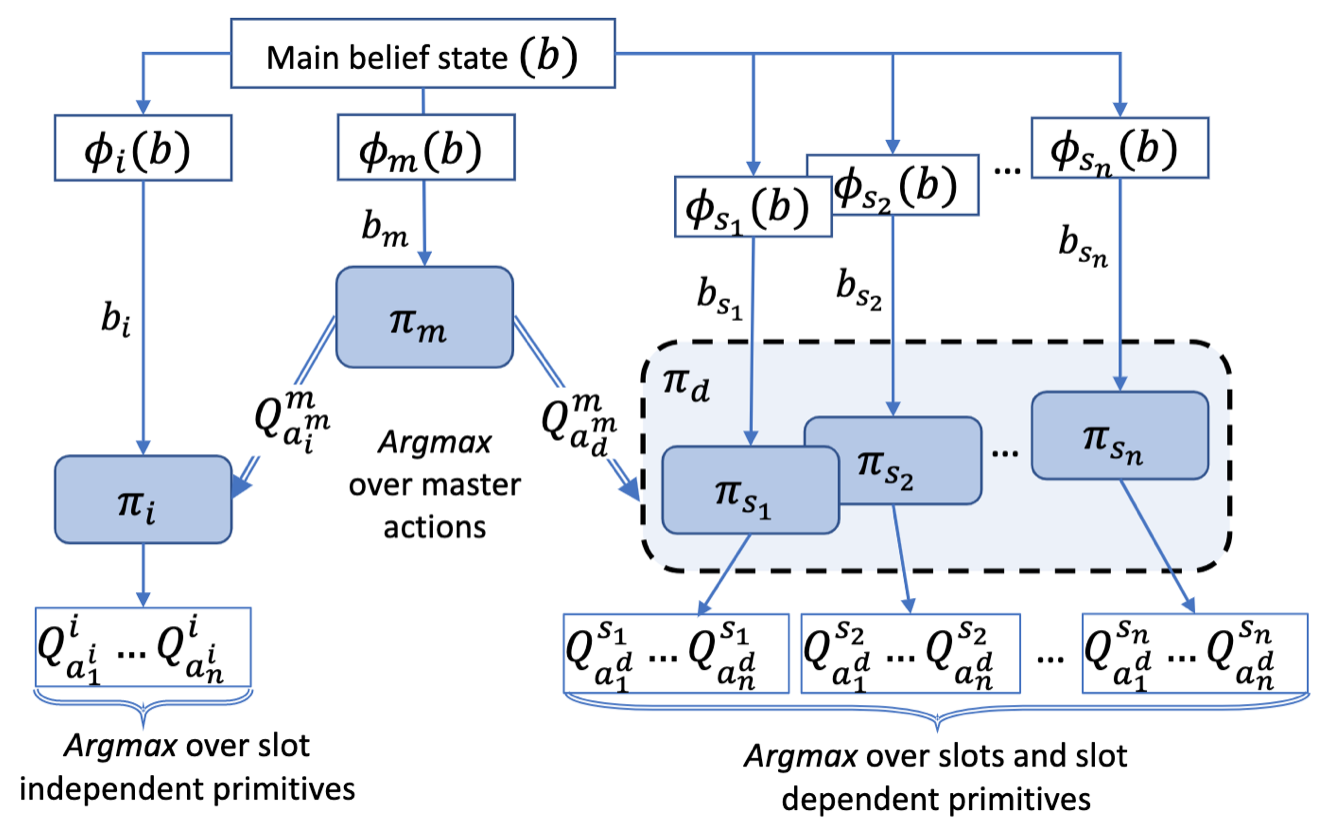}
    \caption{Application of FRL in a task-oriented dialog system}
    \label{fig:17}
\end{figure*}

Cambridge scholars applied FRL\cite{casanueva2018feudal} to task dialog systems for the first time to divide the action space by its relevance to the slots. With this done, only the natural structure of the action space is used, and additional expert knowledge is not required. They put forward a feudal policy structure shown in Figure \ref{fig:17}. The decision-making process for this structure is divided into two steps:
\begin{enumerate}
    \item Determine whether the next action requires slots as parameters.
    \item Select the low-level policy and next action for the corresponding slot based on the decision of the first step.
\end{enumerate}

In general, both HRL and FRL divide the high-dimensional complex action space in different ways to address the low training efficiency of traditional RL methods due to large action space dimensions. HRL divides tasks properly in line with human understanding. However, expert knowledge is required to divide a task into sub-tasks. FRL divides complex tasks based on the logical structure of the action and does not consider mutual constraints between sub-tasks.

\subsubsection{ Model-Based RL}

\begin{figure}
    \centering
    \includegraphics[width=0.5\textwidth]{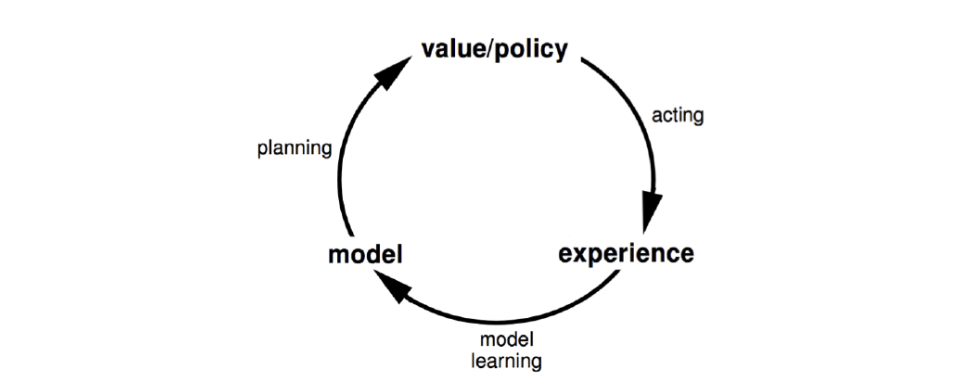}
    \caption{Model-based RL process}
    \label{fig:18}
\end{figure}

The preceding RL methods are model-free. With these methods, a large amount of weakly supervised data is obtained through trial and error interactions with the environment, and then a value network or policy network is trained accordingly. The process is independent of the environment. There is also model-based RL, as shown in Figure \ref{fig:18}. Model-based RL directly models and interacts with the environment to learn a probability transition function of state and reward, namely, an environment model. Then, the system interacts with the environment model to generate more training data. Therefore, model-based RL is more efficient than model-free RL, especially when it is costly to interact with the environment. However, the resulting performance depends on the quality of environment modeling.

Using model-based RL to improve training efficiency is currently an active field of research. Microsoft first applied the classic Deep Dyna-Q (DDQ) algorithm in dialogs\cite{peng2018deep}, as shown by the figure (c) in Figure 19. Before DDQ training starts, we use a small amount of existing dialog data to pre-train the policy model and the world model. Then, we train DDQ by repeating the following steps:
\begin{itemize}
    \item Direct RL: Interact with real users online, update policy models, and store dialog data.
    \item World model training: Update the world model based on collected real dialog data.
    \item Planning: Use the dialog data obtained from interaction with the world model to train the policy model.
\end{itemize}

The world model (as shown in Figure 20) is a neural network that models the probability of environment state transition and rewards. The inputs are the current dialog state and system action. The outputs are the next user action, environment rewards, and dialog termination variables. The world model reduces the human-machine interaction data required by DDQ for online RL (as shown in figure (a) of Figure 19) and avoids ineffective interactions with user simulators (as shown in figure (b) of Figure 19).

\begin{figure*}
    \centering
    \includegraphics[width=1.0\textwidth]{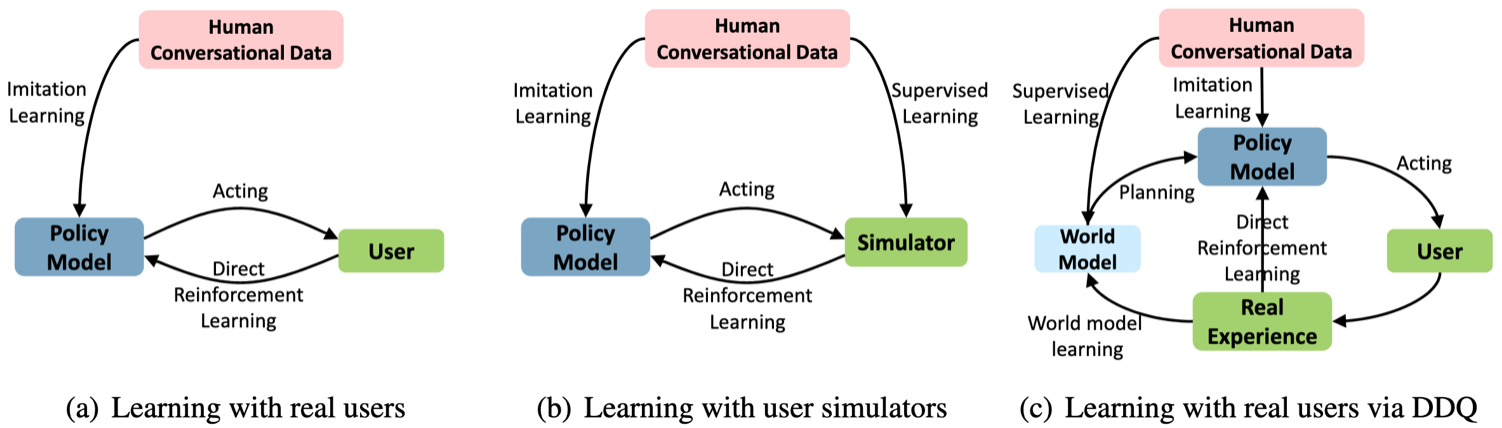}
    \caption{Three RL architectures}
    \label{fig:19}
\end{figure*}

\begin{figure}
    \centering
    \includegraphics[width=0.4\textwidth]{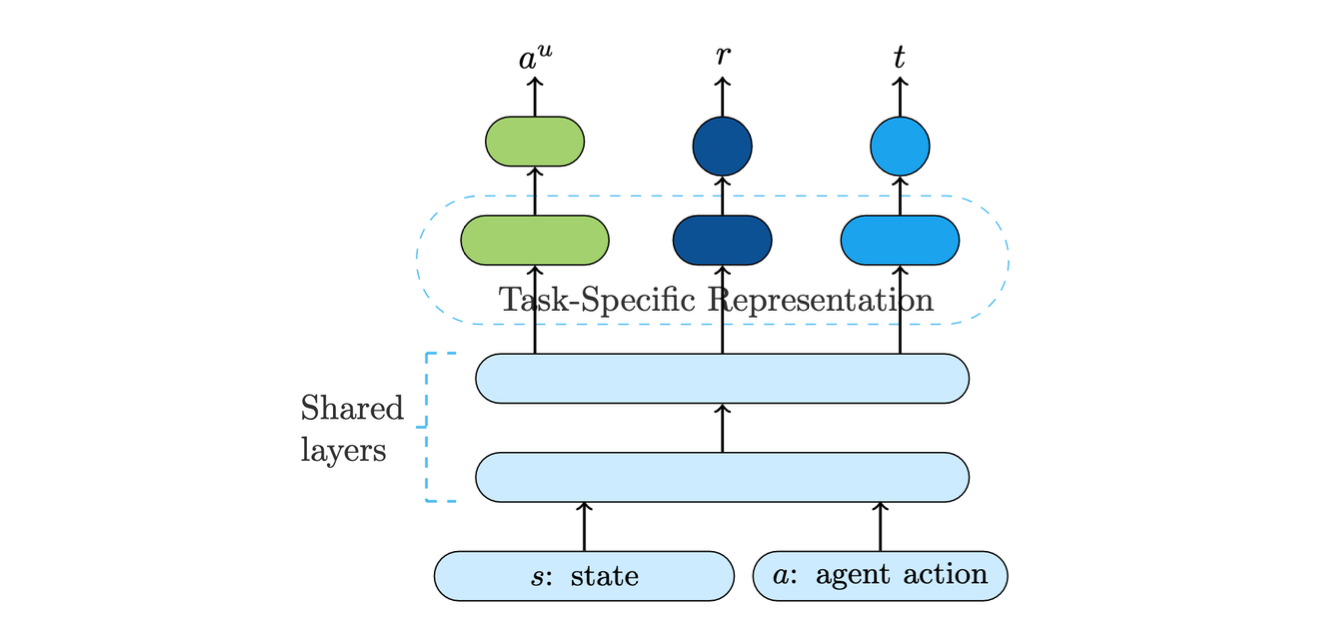}
    \caption{Structure of the world model}
    \label{fig:20}
\end{figure}

Similar to the user simulator in the dialog field, the world model can simulate real user actions and interact with the system's DM. However, the user simulator is essentially an external environment and is used to simulate real users, while the world model is an internal model of the system.

Microsoft researchers have made improvements based on DDQ. To improve the authenticity of the dialog data generated by the world model, they proposed\cite{su2018discriminative} to improve the quality of the generated dialog data through adversarial training. Considering when to use the data generated through interaction with the real environment and when to use data generated through interaction with the world model, they discussed feasible solutions in a paper\cite{wu2019switch}. They also discussed a unified dialog framework to include interaction with real users in another paper\cite{zhang2019budgeted}. This human-teaching concept has attracted attention in the industry as it can help in the building of DMs. This will be further explained in the following sections.

\subsubsection{Human-in-the-Loop}

We hope to make full use of human knowledge and experience to generate high-quality data and improve the efficiency of model training. Human-in-the-loop RL\cite{abel2017agent}is a method to introduce human beings into robot training. Through designed human-machine interaction methods, humans can efficiently guide the training of RL models. To further improve the training efficiency of the task dialog systems, researchers are working to design an effective human-in-the-loop method based on the dialog features.

\begin{figure}
    \centering
    \includegraphics[width=0.4\textwidth]{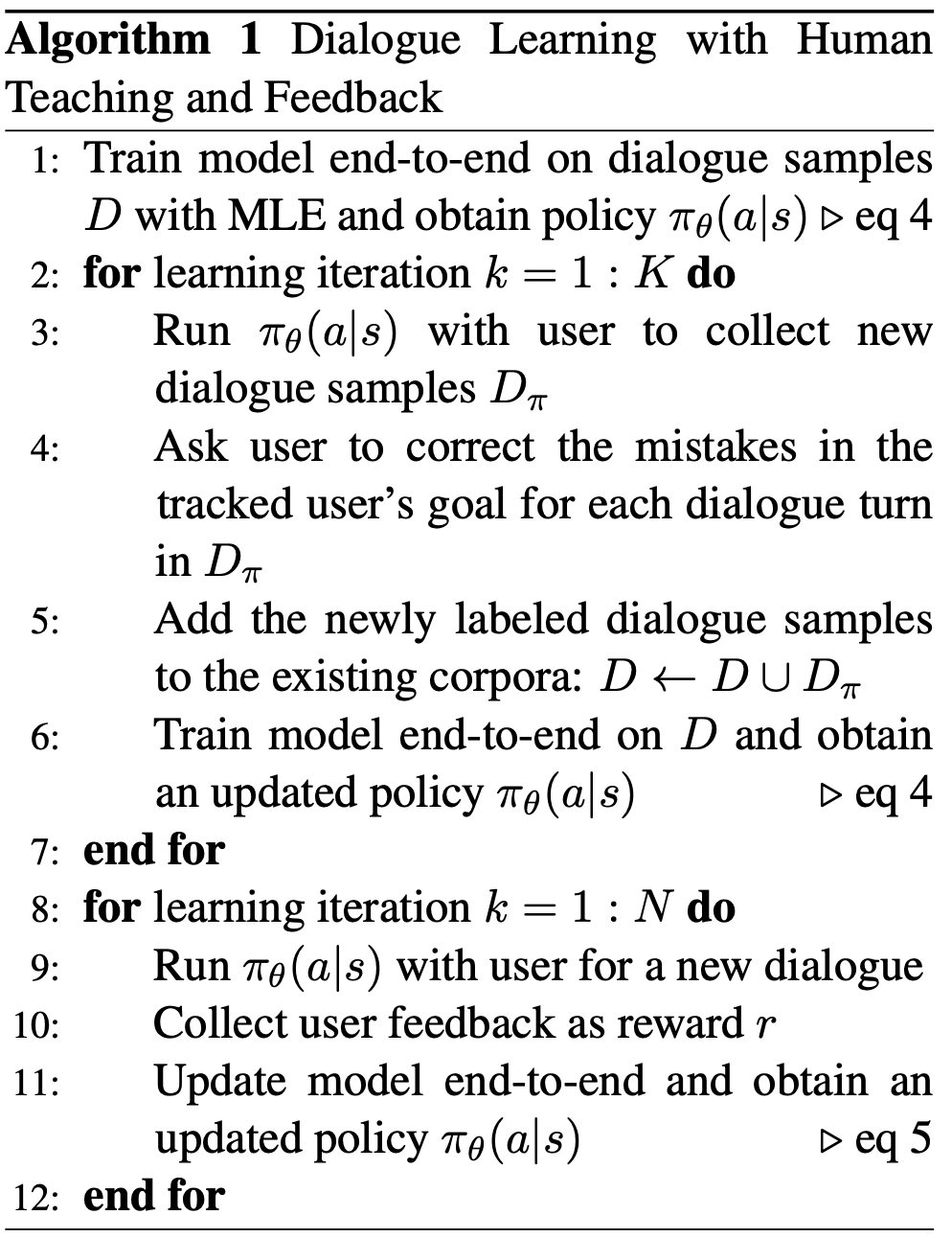}
    \caption{Composite learning combining supervised pre-training, imitation learning, and online RL}
    \label{fig:21}
\end{figure}

Google researchers proposed a composite learning method combining human teaching and RL\cite{abel2017agent} (as shown in Figure 21), which adds a human teaching stage between supervised pre-training and online RL, allowing humans to tag data to avoid the covariate shift caused by supervised pre-training\cite{ross2011reduction}. Amazon researchers also proposed a similar human teaching framework\cite{abel2017agent}: In each round of dialog, the system recommends four responses to the customer service expert. The customer service expert determines whether to select one of these responses or create a new response. Finally, the customer service expert sends the selected or created response to the user. With this method, developers can quickly update the capabilities of the dialog system.

In the preceding method, the system passively receives the data tagged by humans. However, a good system should actively ask questions and seek help from humans. One paper\cite{chen2017agent} introduced the companion learning architecture (as shown in Figure 22), which adds the role of a teacher (human) to the traditional RL framework. The teacher can correct the responses of the dialog system (the student, represented by the switch on the left side of the figure) and evaluate the student's response in the form of intrinsic reward (the switch on the right side of the figure). For the implementation of active learning, the authors put forward the concept of dialog decision certainty. The student policy network is sampled multiple times through dropout to obtain the estimated approximate maximum probability of the desired action. Then the moving average of several dialog rounds is calculated through the maximum probability and used as the decision certainty of the student policy network. If the calculated certainty is lower than the target value, the system determines whether a teacher is required to correct errors and provide reward functions based on the difference between the calculated decision certainty and the target value. If the calculated certainty is higher than the target value, the system stops learning from the teacher and makes judgments on its own.

\begin{figure*}
    \centering
    \includegraphics[width=0.8\textwidth]{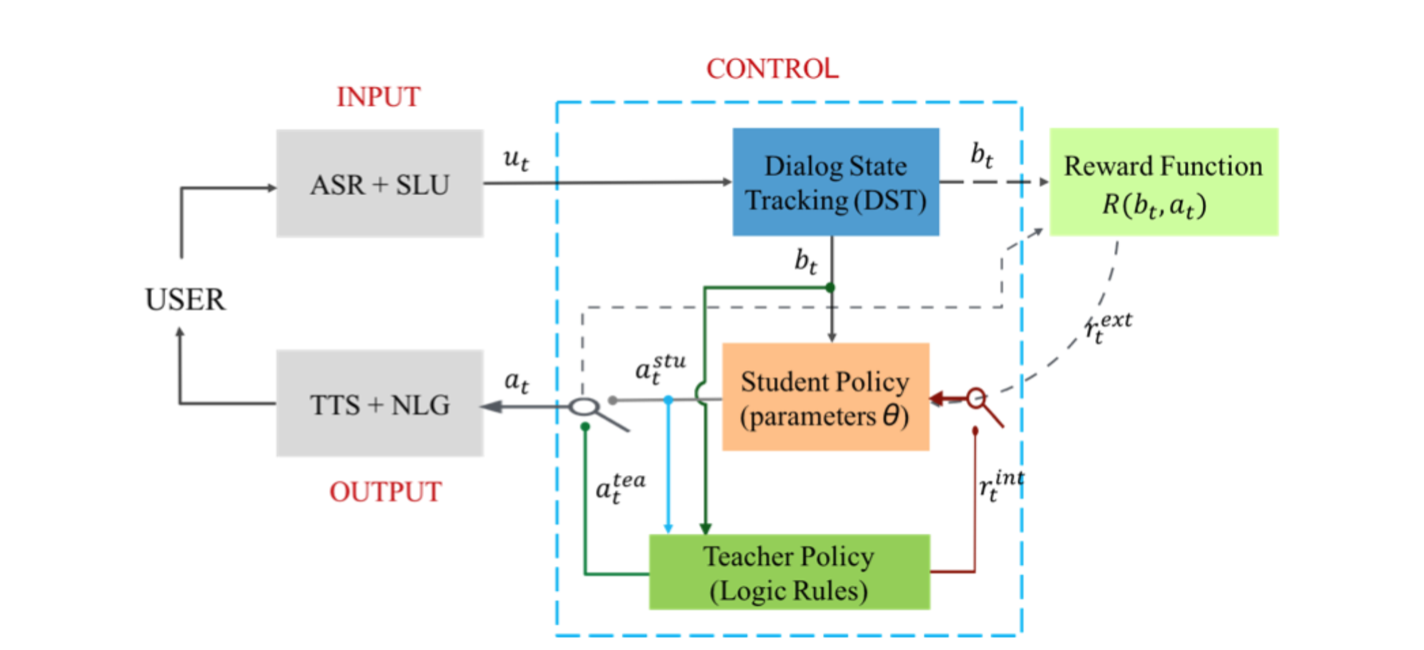}
    \caption{The teacher corrects the student's response (on the left) or evaluates the student's response (on the right).}
    \label{fig:22}
\end{figure*}

The key to active learning is to estimate the certainty of the dialog system regarding its own decisions. In addition to dropping out policy networks, other methods include using hidden variables as condition variables to calculate the Jensen-Shannon divergence of policy networks\cite{wang2019incremental} and making judgments based on the dialog success rate of the current system\cite{zhang2019budgeted}.

\section{Dialog Management Framework of the Intelligent Robot Conversational AI team}

To ensure stability and interpretability, the industry primarily uses rule-based DM models. The Intelligent Robot Conversational AI Team at Alibaba's DAMO Academy began to explore DM models last year. When building a real dialog system, we need to solve two problems: (1) how to obtain a large amount of dialog data in a specific scenario and (2) how to use algorithms to maximize the value of data.

Currently, we plan to complete the model framework design in four steps, as shown in Figure 23.

\begin{figure*}
    \centering
    \includegraphics[width=0.95\textwidth]{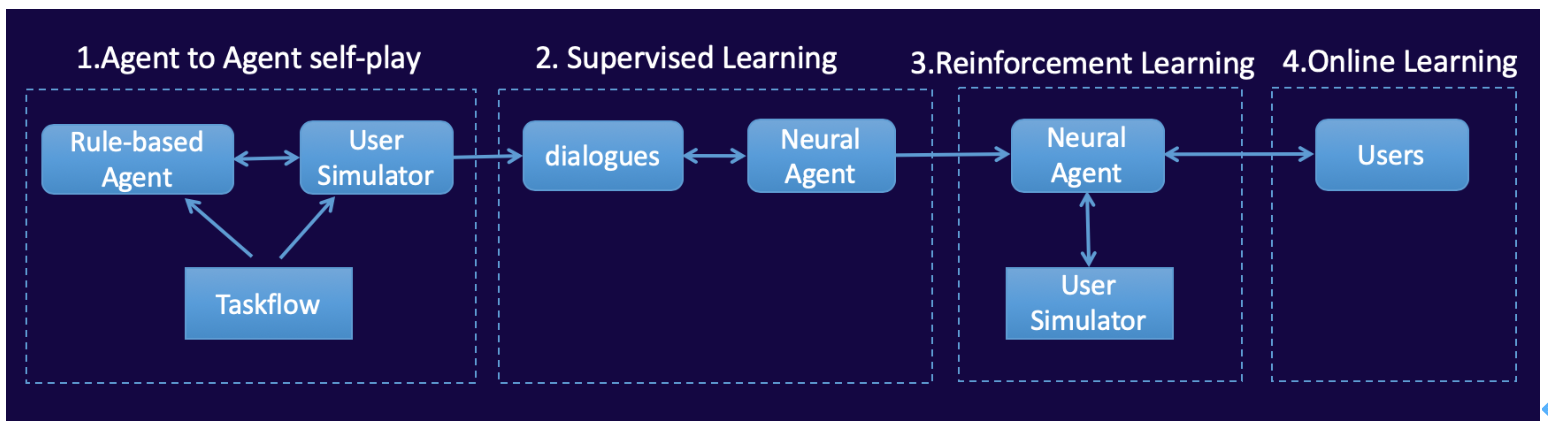}
    \caption{Four steps of DM model design}
    \label{fig:23}
\end{figure*}

\textbf{Step 1:} First, use the dialog studio independently developed by the Intelligent Robot Conversational AI team to quickly build a dialog engine called TaskFlow based on rule-based dialog flows and build a user simulator with similar dialog flows. Then, have the user simulator and TaskFlow continuously interact with each other to generate a large amount of dialog data.

\textbf{Step 2:} Train a neural network through supervised learning to build a preliminary DM model that has capabilities basically equivalent to a rule-based dialog engine. The model can be expanded by combining semantic similarity matching and end-to-end generation. Dialog tasks with a large action space are divided using the HRL method.

\textbf{Step 3:} In the development phase, make the system interact with an improved user simulator or AI trainers and continuously enhance the system dialog capability based on off-policy ACER RL algorithms.

\textbf{Step 4:} After the human-machine interaction experience is verified, launch the system and introduce human roles to collect real user interaction data. In addition, use some UI designs to easily introduce user feedback to continuously update and enhance the model. The obtained human-machine dialog data will be further analyzed and mined for customer insight.

At present, the RL-based DM model we developed can complete 80\% of the dialog with the user simulator for moderately complex dialog tasks, such as booking a meeting room, as shown in Figure 24.
\begin{figure*}
    \centering
    \includegraphics[width=0.8\textwidth]{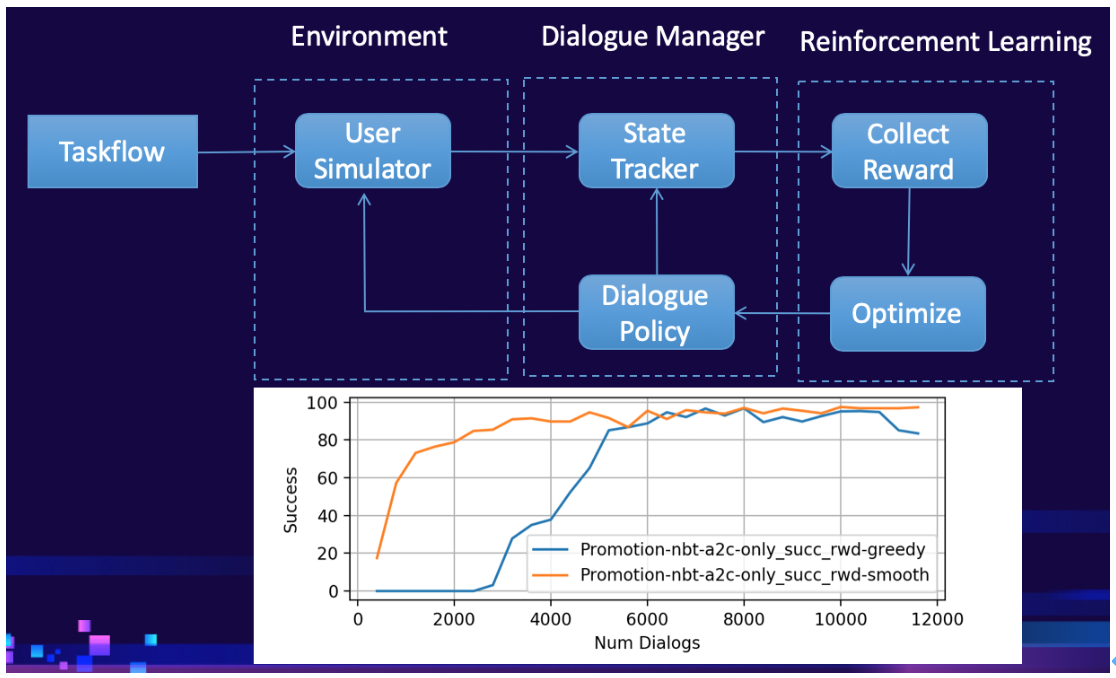}
    \caption{Framework and evaluation indicators of the DM model developed by the Intelligent Robot Conversational AI team}
    \label{fig:my_label}
\end{figure*}

\section{Summary}

This article provides a detailed introduction of the latest research on DM models, focusing on three shortcomings of traditional DM models: 
\begin{itemize}
    \item Poor scalability
    \item Insufficient tagged data
    \item Low training efficiency
\end{itemize}

To address scalability, common methods for processing changes in user intents, dialog bodies, and the system action space include semantic similarity matching, knowledge distillation, and sequence generation. To address insufficient tagged data, methods include automatic machine tagging, effective dialog structure mining, and efficient data collection policies. To address the low training efficiency of traditional DM models, methods such as HRL and FRL are used to divide action spaces into different layers. Model-based RL methods are also used to model the environment and improve training efficiency. Introducing human-in-the-loop into the dialog system training framework is also a current focus of research. Finally, we discussed the current progress of the DM model developed by the Intelligent Robot Conversational AI team of Alibaba's DAMO Academy. We hope this summary can provide some new insights to support your own research on DM.

\section*{Acknowledgement}
The authors would like to thank Alibaba Cloud International Team for their efforts on the translation.

\bibliography{reference}
\bibliographystyle{acl_natbib}

\end{document}